\documentclass{article}
\usepackage{spconf,amsmath,graphicx}
\usepackage{booktabs}
\usepackage{amsthm}

\usepackage{graphicx}
\usepackage{graphics}
\usepackage{lineno,hyperref}
\usepackage{array}
\usepackage{amsmath,amssymb}
\usepackage{mathrsfs}
\usepackage{subfigure} 
\usepackage{multirow}
\usepackage{algorithm}
\usepackage{algorithmic}
\usepackage{colortbl}
\usepackage{amssymb,subfigure,cases}
\title{BiP-Net: Bidirectional Perspective Strategy based Arbitrary-Shaped Text Detection Network}
\name{Chuang Yang, Mulin Chen, Yuan Yuan, Qi Wang$^{*}$\thanks{$^{*}$Qi Wang is the corresponding author.}}
\address{School of Computer Science and Artificial Intelligence, Optics and Electronics (iOPEN), \\Northwestern Polytechnical University, Xi'an 710072, Shaanxi, P. R. China}

\begin{document}
%
\maketitle

\begin{abstract}
Detecting irregular-shaped text instances is the main challenge for text detection. Existing approaches can be roughly divided into top-down and bottom-up perspective methods. The former encodes text contours into unified units, which always fails to fit highly curved text contours. The latter represents text instances by a number of local units, where the complicated network and post-processing lead to slow detection speed. In this paper, to detect arbitrary-shaped text instances with high detection accuracy and speed simultaneously, we propose a \textbf{Bi}directional \textbf{P}erspective strategy based \textbf{Net}work (BiP-Net). Specifically, a new text representation strategy is proposed to represent text contours from a top-down perspective, which can fit highly curved text contours effectively. Moreover, a contour connecting (CC) algorithm is proposed to avoid the information loss of text contours by rebuilding interval contours from a bottom-up perspective. The experimental results on MSRA-TD500, CTW1500, and ICDAR2015 datasets demonstrate the superiority of BiP-Net against several state-of-the-art methods.
\end{abstract} 
\section{Introduction}
\label{intro}
Scene text detection~\cite{DBLP:conf/icassp/0010LXY0SBS20,DBLP:conf/icassp/CaoZ20,DBLP:conf/icassp/ZhongJH17}, which aims to detect text instances with high detection accuracy and speed simultaneously, is greatly improved by detection~\cite{wang2020multitask,DBLP:journals/tip/YuanXW19,DBLP:conf/aaai/LiCNW17} and segmentation~\cite{DBLP:journals/tip/WangGL19,DBLP:journals/tomccap/YuanFLF19} technologies. According to the differences of text representation strategies, existing text detection methods can be roughly divided into top-down perspective methods~\cite{zhou2017east,wang2020textray,wang2020contournet} and bottom-up perspective methods~\cite{zhu2021fourier,tang2019seglink++,baek2019character}. 

Though top-down perspective methods enjoy fast detection speed, they usually can not fit some irregular-shaped text contours accurately (such as highly curved text contours). For example, Wang $et~al.$~\cite{wang2020textray} sampled each text contour point in one direction. However, for the highly curved text contours, multiple contour points may reside in the same direction, which leads to the failure of contour reconstruction. 

For bottom-up perspective methods, they can fit any geometries accurately. However, the complicated network and post-processing lead to a huge number of parameters and low detection speed. For example, Ma~ $et~al.$~\cite{baek2019character} combined a series of characters and the corresponding affinities by a complicated pipeline to reconstruct text contours, which brings the expensive computational cost to the inference process.
\begin{figure}
	\centering
	\subfigure[Sampling contour points from a top-down perspective.]{
		\begin{minipage}[b]{0.75\linewidth}
			\includegraphics[width=1\linewidth]{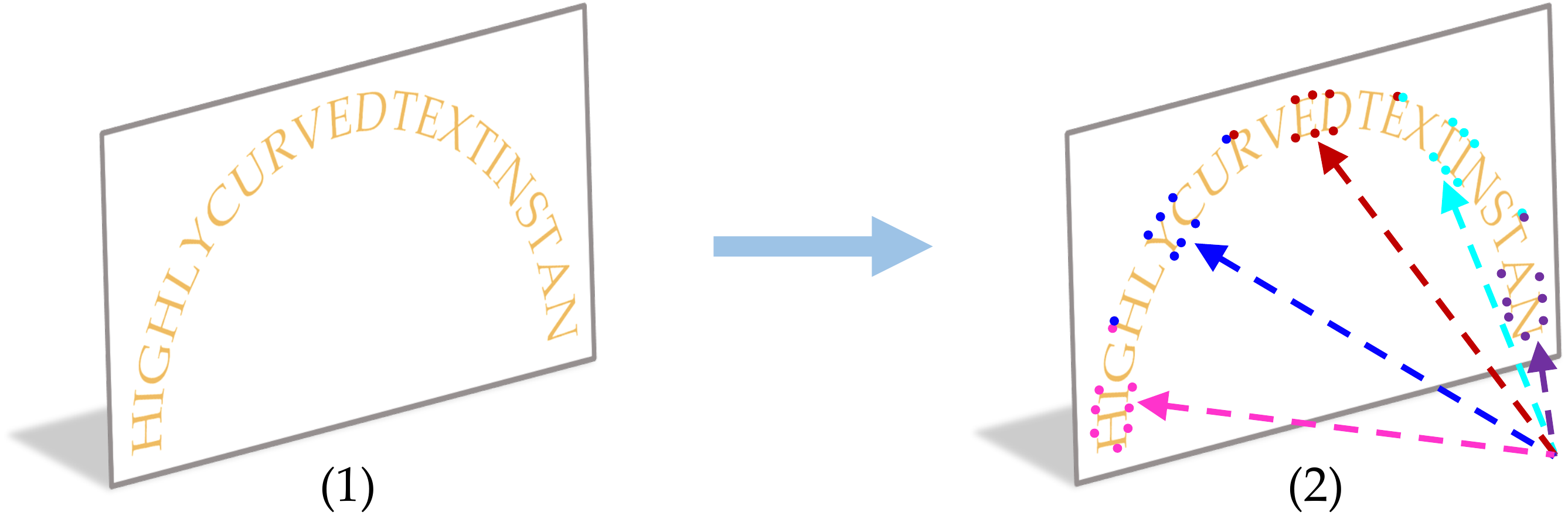}
	\end{minipage}}
	\vspace{-0.3cm}
	
	\subfigure[Rebuilding text contour from a bottom-up perspective.]{
		\begin{minipage}[b]{0.93\linewidth}
			\includegraphics[width=1\linewidth]{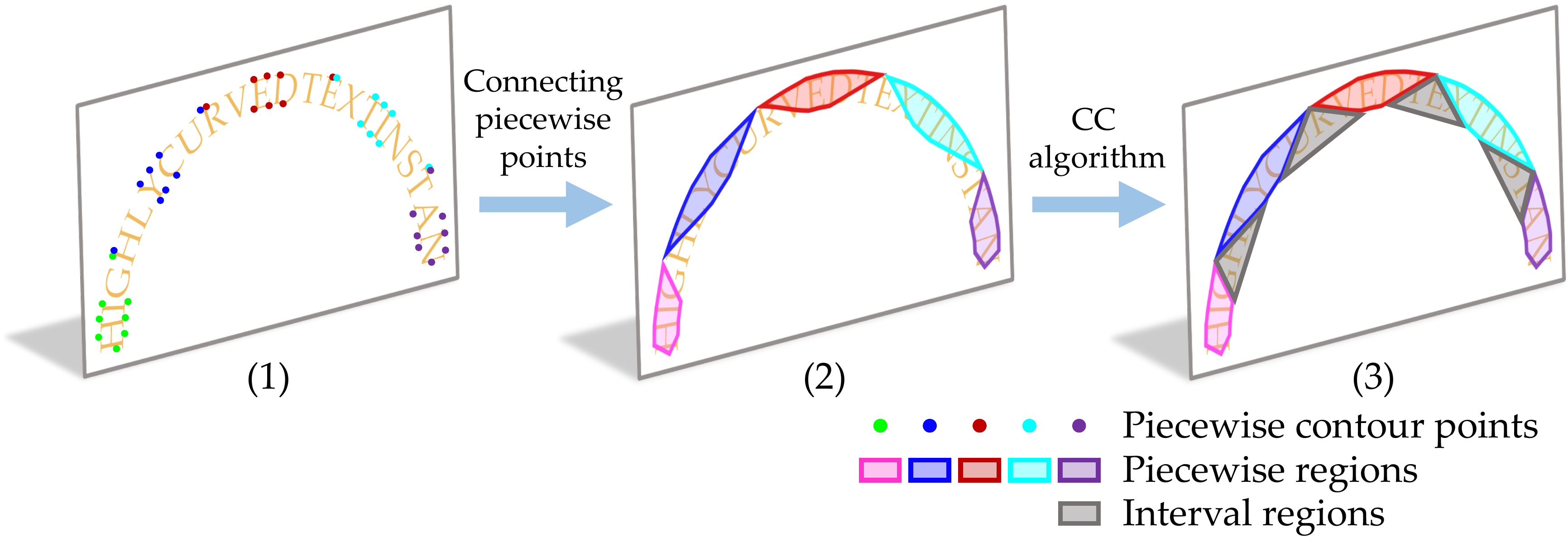}
	\end{minipage}}
	\vspace{-0.3cm}
	\caption{Demonstration of bidirectional perspective strategy.}
	\label{V1}
	\vspace{-0.5cm}
\end{figure}

Considering the aforementioned limitations, we propose a Bidirectional Perspective strategy based Network (BiP-Net) to detect arbitrary-shaped text with high detection accuracy and speed simultaneously. It represents text contours from a top-down perspective and avoids the information loss of text contours by rebuilding interval contours from a bottom-up perspective. Specifically, BiP-Net represents text contours by multiple ray clusters (introduced in Section~\ref{representing}) and samples different piecewise contour point sets from different ray clusters (as shown in Fig.~\ref{V1}~(a)~(2)). Since only the points from the same piecewise contour point set are ordered and can be connected in sequence directly, to rebuild text contour, multiple piecewise contours are generated by connecting the points from different piecewise contour point sets in sequence at first (as shown in Fig.~\ref{V1}~(b)~(2)). Then, a Contour Connecting (CC) algorithm is proposed to generate interval contours between adjacent piecewise contours to avoid the information loss of rebuilt highly curved text contours (as shown in Fig.~\ref{V1}~(b)~(3)). 

The main contributions of this paper are as follows:
\begin{itemize}
\item[$\bullet$] A new top-down perspective text representation strategy is proposed to fit text contours through multiple ray clusters. It can represent arbitrary-shaped text contours more effectively compared with previous methods.

\item[$\bullet$] A bottom-up perspective Contour Connecting (CC) algorithm is proposed to generate interval region contours, which avoids the information loss of rebuilt highly curved text contours effectively.

\item[$\bullet$] A bidirectional perspective text detection framework is constructed. It achieves the best balance between detection accuracy and speed compared with several recent state-of-the-art approaches.
\end{itemize}
\section{RELATED WORK}
\label{related}

Recent numerous CNN-based text detection methods have been proposed, which can be roughly classified into top-down perspective methods and bottom-up perspective methods.

Top-down perspective methods treat text instances as unified units. Liao~$et~al$.~\cite{liao2018rotation} adopted rotation-sensitive features to adapt multi-oriented text instances.
Zhang~$et~al$.~\cite{DBLP:conf/icassp/0010LXY0SBS20} improved the ability to distinguish positive and false samples through self-attention mechanism. Zhou~$et~al$.~\cite{zhou2017east} predicted dense boxes to detect text instances without anchors. Wang~$et~al$.~\cite{wang2020textray} represented text contours through a fixed number of text contour points. However, they could not fit highly-curved text contours. Bottom-up perspective methods decompose text instances into multiple local units. Wang~$et~al$.~\cite{wang2020contournet} predicted dense text contour points to fit text contours. Zhu~$et~al$.~\cite{zhu2021fourier} modeled text contours through multiple Fourier signature vectors. Some works ~\cite{tang2019seglink++,baek2019character,long2018textsnake,feng2019textdragon} predicted character-level text patches and combined them to generate integrity text instances. Three researches~\cite{tian2019learning,xu2019textfield,zhang2020opmp} segmented text regions directly and distinguished adjacent text instances by predicting their affinities. Law~$et~al$.~\cite{law2018cornernet} decomposed text into different sensitive regions by corner points and combined those regions to obtain the integrity text region. Wang~$et~al$.~\cite{wang2019efficient} roughly located text instances through shrink masks and rebuild text contours by extending them to text masks. Although bottom-up perspective methods can accurately fit any geometries, the complicated framework influences the detection speed seriously.

\begin{figure}
	\begin{center}
		\includegraphics[width=0.425\textwidth]{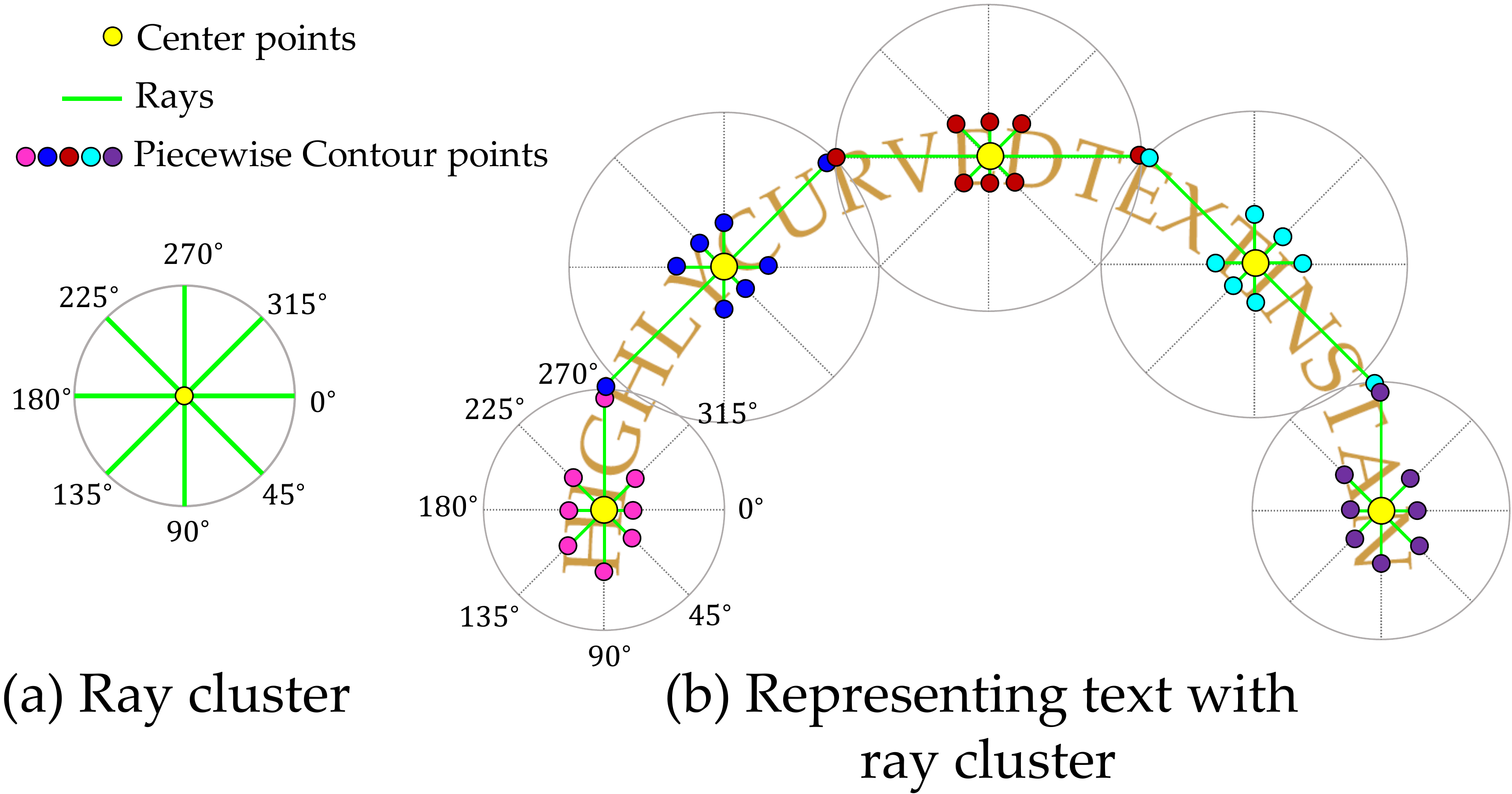}
	\end{center}
	\vspace{-0.7cm}
	\caption{Illustration of the ray cluster based text representation method.}
	\label{V3}
	\vspace{-0.31cm}
\end{figure}

\section{PROPOSED METHOD}
\label{method}
In this section, the proposed novel text representation strategy is explained in Section~\ref{representing} at first. Then, the overall architecture of the network is illustrated in Section~\ref{architecture}. Next, we describe the Contour Connecting (CC) algorithm in Section~\ref{CC}. In the end, Section~\ref{label} shows the generation process of ground-truth and Section~\ref{loss} gives the loss function.

\subsection{Top-Down Perspective Text Representation Strategy}
\label{representing}
To effectively and efficiently fit arbitrary-shaped text contours, a new top-down perspective text representation strategy is proposed. It fits text contours through multiple ray clusters. As we can see from Fig.~\ref{V3}~(a), one ray cluster consists of a center point and multiple rays. The rays are defined as vectors that evenly emit from the center point to text contour in multiple directions and they consist of ray distances and ray directions. For one text instance, as shown in Fig.~\ref{V3}~(b), it is represented through $N$ ray clusters. For one ray cluster, which consists of a center point and $M$ rays. $N$ and $M$ are set to 5 and 8 in this paper.

\begin{figure}[h]
	\begin{center}
		\includegraphics[width=0.47\textwidth]{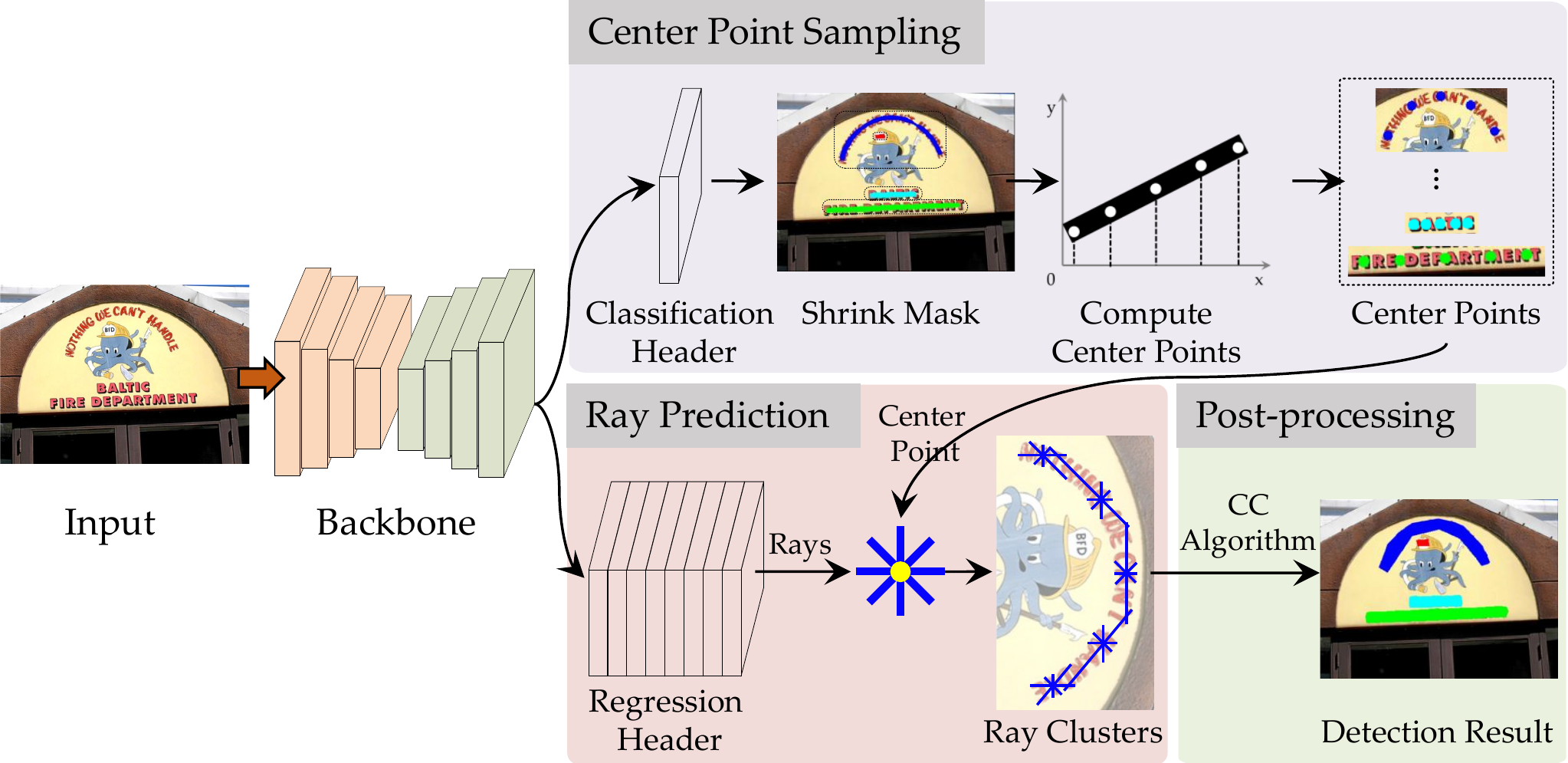}
	\end{center}
	\vspace{-0.5cm}
	\caption{The overall architecture of the proposed BiP-Net.}
	\label{V2}
\end{figure}

\subsection{Overall Architecture}
\label{architecture}
As discussed before, existing methods either can not effectively fit irregular-shaped text contours or the complicated pipeline leads to low detection speed. To detect arbitrary-shaped text instances with high detection accuracy and speed simultaneously, we construct BiP-Net based on the proposed text representation strategy and CC algorithm.

BiP-Net is a novel single-shot anchor-free text detector (as we can see from Fig.~\ref{V2}), which consists of backbone, center point sampling module, ray prediction module, and post-processing. The backbone is composed of ResNet and Feature Pyramid Network (FPN), which is used to extract concatenated feature map possesses  multi-scale context information. For the center point sampling module and ray prediction module, they conduct on the concatenated feature map to predict contour point coordinates. Specifically, the center point sampling module segments shrink masks~\cite{zhou2017east} by classification header that consists of one 1$\times$1 convolutional layer. Then, the center point coordinates are computed. Their abscissas are obtained by an equidistant sampling of the shrink mask on the X-axis, and the median values of the shrink mask on the Y-axis corresponding to the abscissas are taken as ordinates. For ray prediction module, it regresses ray distances in different directions by regression header that consists of 8 1$\times$1 convolutional layers. Ray clusters are formed by combining the center point and the corresponding rays (such as the Fig.~\ref{V3}(a)) and text contour points are obtained by sampling the endpoints of the ray clusters. After getting text contour points, the text contour can be generated through post-processing.

\subsection{Contour Connecting (CC) Algorithm}
\label{CC}
Since the interval regions are ignored when treating all piecewise contours as predicted text contour directly (Fig.~\ref{V1}~(b)~(2)), which deeply influences text recognition, CC algorithm is proposed to detect interval regions to improve the integrity of detection results.

\begin{figure}
	\begin{center}
		\includegraphics[width=0.42\textwidth]{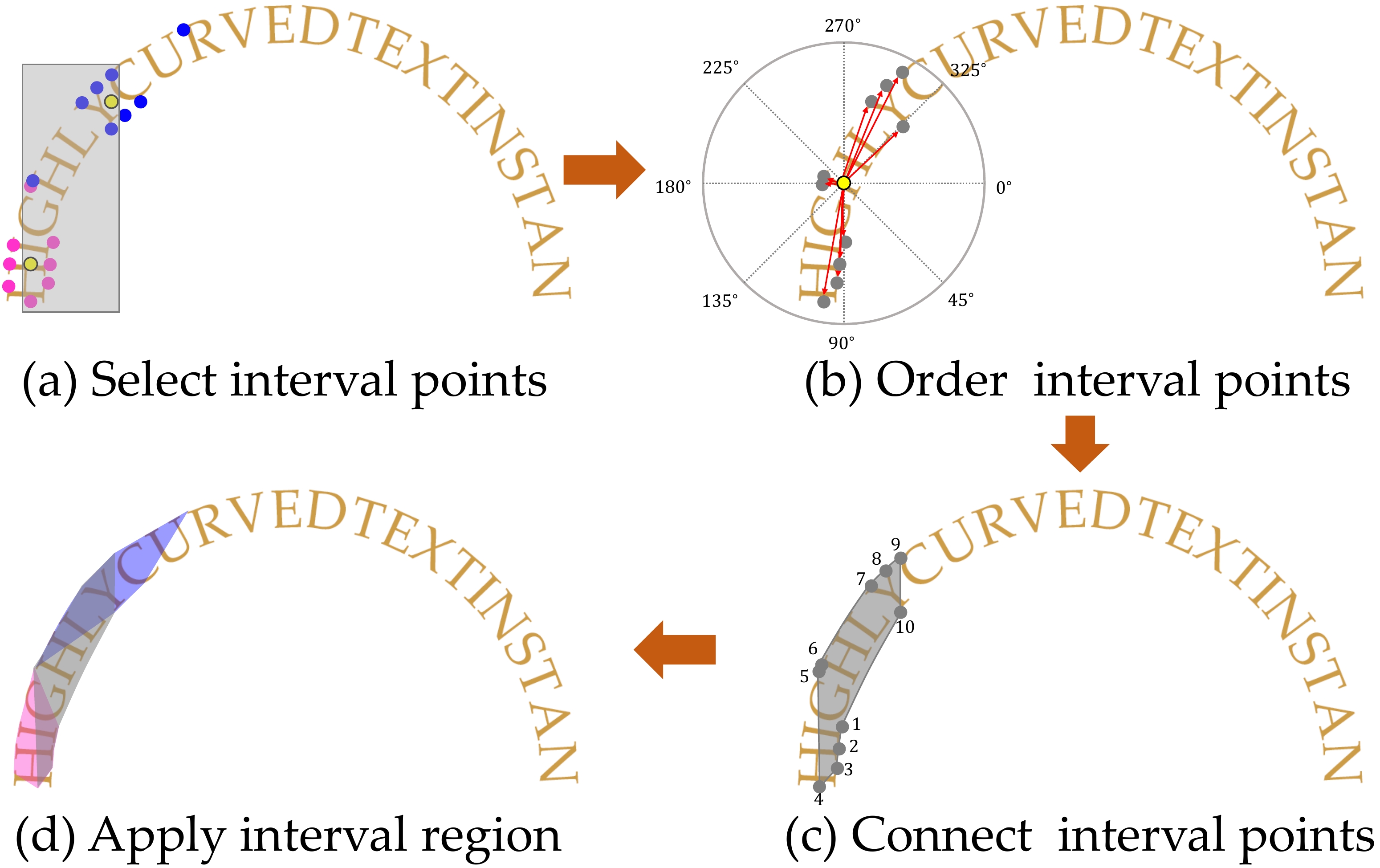}
	\end{center}
	\vspace{-0.5cm}
	\caption{Visualization the process of CC algorithm.}
	\label{V7}
	\vspace{-0.3cm}
\end{figure}

The algorithm selects the points between the center points of adjacent piecewise contour points as interval region contour points (Fig.~\ref{V7}~(a)). Since they are unordered, to connect them in sequence, the algorithm need to order them through the following steps: (1) compute the center point of interval region contour points through the way to sample center point from shrink mask (Section~\ref{architecture}); (2) all interval region contour points are sorted in descending order through the angles between the contour points and center point (Fig.~\ref{V7}~(b)); (3) the sorting points (Fig.~\ref{V7}~(c)) are connected in sequence to generate the interval region; (4) the region is combined with adjacent piecewise regions (Fig.~\ref{V7}~(d)). The final integrity text contour is obtained by performing the above process for all adjacent piecewise contours (Fig.~\ref{V1}~(b)~(3)).

\subsection{Objective Function}
\label{loss}
As mentioned before, the proposed BiP-Net is optimized under the supervision of shrink mask and ray distances. ${\cal L}_{sm}$ and ${\cal L}_{ray}$ are adopted to evaluate the predicted shrink mask and ray distances respectively.
The formulas of the two kinds of loss function are defined as:
\vspace{-0.15cm}
\begin{equation}
{\cal L}_{sm}=1-\frac{2\times \left| P\cap G \right|+1 }{\left| P \right|+\left| G \right|+1 },
\vspace{-0.7cm}
\end{equation}
\begin{equation} 
\begin{gathered}
{\cal L}_{ray}=\frac{1}{N}\sum_{j=1}^{N}{({\sum_{i=1}^{M}{\left\{ \frac{\log \left( \frac{d_{P_i}}{d_{G_i}} \right) \,\,,if\,\,d_{P_i}>=d_{G_i}}{\log \left( \frac{d_{G_i}}{d_{P_i}} \right) \,\,, else\,\,\,\,\,\,\,\,\,\,\,\,\,\,\,\,\,\,\,\,\,\,\,\,\,\,\,} \right.})_j}},
\vspace{-0.2cm}
\end{gathered}
\end{equation}          
where $P$, $G$ denote the predicted and ground-truth shrink masks respectively. $N$, $M$ are the numbers of piecewise contours and ray distances. $d_{P_i}$ and $d_{G_i}$ denote the predicted and ground-truth of ray distance in $i_{th}$ direction.

The network is trained using the following unfied loss:
\vspace{-0.2cm}
\begin{equation}
{\cal L}={\cal L}_{sm}+\lambda {\cal L}_{ray},
\vspace{-0.2cm}
\end{equation}
where $\lambda$ is weighting weight that is set to 0.25.

\subsection{Ground Truth Generation}
\label{label}
In this paper, we use two kinds of ground truth. The shrink mask ground-truth is generated by:
\vspace{-0.2cm}
\begin{equation}
M_{GT}\left( x_i \right) =\left\{ \begin{array}{c}
1, if\,\,x_i\subset text\,\,~~~~~~~~~~~~\\
0, if\,\,x_i\subset background\\
\end{array} \right.,
\vspace{-0.2cm}
\end{equation}
where $M_{GT}\left( x_i \right)$ is the value of $i$th pixel in ground-truth. A binary shrink mask map ground truth $M_{GT}$ is generated through the formula above.

The ray distance ground-truth $D_{GT}$ is defined as the Euclidean distance between contour point and center point:
\begin{equation}
D_{GT}\left( p_{ct},p_{cp} \right) =||p_{ct}-p_{cp}||^2_2,
\vspace{-0.2cm}
\end{equation}
where $p_{ct}$ and $p_{cp}$ are the coordinates of the contour point and center point respectively.

\section{EXPERIMENTAL RESULTS}
\label{experiment}
Three popular scene text detection datasets are used to evaluate the performance of our method. SynthText~\cite{gupta2016synthetic} is used to pre-train our model. MSRA-TD500~\cite{yao2012detecting} consists of long multi-lingual text instances. It contains 700 training images and 200 testing images, where the training set includes 400 images from HUST-TR400~\cite{yao2014unified}. ICDAR2015~\cite{karatzas2015icdar} contains 1000 training images and 500 testing images, and consists of multi-oriented text instances in complex background. CTW-1500~\cite{yuliang2017detecting} is a dataset that mainly consists of curved text instances, which contains 1000 images for training and 500 images for testing. To make a fair comparison, we use the F-measure and frames per second (FPS) to evaluate the detection accuracy and speed respectively.
\begin{table}[]
	\footnotesize
	\setlength{\tabcolsep}{1.5mm}
	\caption{Comparison with other state-of-the-art methods on MSRA-TD500, CTW1500, and ICDAR2015. ``Ext.'' denotes extra training data. ``\textcolor{red}{Red}'' denotes the best results.}
	\label{comparison}
	\centering
	\renewcommand\arraystretch{0.8}
	\begin{tabular}{c|c|cccc}
		\toprule
		Methods    & \multicolumn{1}{c|}{Ext.} & Precision & Recall  & F-measure & FPS  \\ \midrule
		&                           & \multicolumn{4}{c}{MSRA-TD500}       \\ \midrule       
		EAST~\cite{zhou2017east}        &   $\times$     & 87.3      & 67.4   & 76.1      & 13.2 \\  
		PAN~\cite{wang2019efficient}        &   $\times$     & 80.7      & 77.3   & 78.9      & 30.2 \\                                     
		RRD~\cite{liao2018rotation}        &   \checkmark   & 87.0      & 73.0   & 79.0      & 10.0   \\ 
		CRAFT~\cite{baek2019character}      &   \checkmark   & 88.2      & 78.2   & 82.9      & 8.6  \\ 
		SAE~\cite{tian2019learning}        &   \checkmark     & 84.2      & 81.7   & 82.9      & -- \\ 
		\textbf{Ours}       &   \checkmark     & 91.6      & 81.8   & \textbf{\textcolor{red}{86.4}}      &  \textbf{\textcolor{red}{30.7}} \\ \midrule
		&                           & \multicolumn{4}{c}{CTW1500}       \\ \midrule
		TextSnake~\cite{long2018textsnake}  &   \checkmark   & 67.1      & 85.3   & 75.6    & 1.1  \\ 
		TextRay~\cite{wang2020textray}       &   $\times$   & 82.8      & 80.4   & 81.6      & --  \\  
		OPMP~\cite{zhang2020opmp}       &   \checkmark   & 85.1      & 80.8   & 82.9      & 1.4  \\  
		FCENet~\cite{zhu2021fourier} &   $\times$     & 85.7      & 80.7   & 83.1      & --  \\ 
		ContourNet~\cite{wang2020contournet} &   $\times$     & 83.7      & 84.1   & 83.9      & 4.5  \\
		\textbf{Ours}       &   \checkmark     & 87.7      & 82.6   &  \textbf{\textcolor{red}{85.1}}      &  \textbf{\textcolor{red}{35.6}} \\ \midrule
		&                           & \multicolumn{4}{c}{ICDAR2015}       \\ \midrule
		CornerNet~\cite{law2018cornernet}  &   \checkmark   & 94.1      & 70.7   & 80.7      & 3.6   \\ 
		ESTD~\cite{DBLP:conf/icassp/0010LXY0SBS20}  &   $\times$   & 85.8      & 77.8   & 81.5      & --   \\ 
		SegLink++~\cite{tang2019seglink++}  &   \checkmark   & 83.7      & 80.3   & 82.0        & 7.1  \\ 
		TextField~\cite{xu2019textfield}  &   \checkmark   & 84.3      & 80.5   & 82.4      & 6.0    \\ 
		TextDragon~\cite{feng2019textdragon} &   \checkmark   & 84.8      & 81.8   & 83.1      & --   \\ 
		\textbf{Ours}       &   \checkmark     & 86.9      & 82.1   &  \textbf{\textcolor{red}{83.9}}      &  \textbf{\textcolor{red}{24.8}} \\ \bottomrule
	\end{tabular}
\end{table}

\begin{figure}[h]
	\begin{center}
		\includegraphics[width=0.46\textwidth]{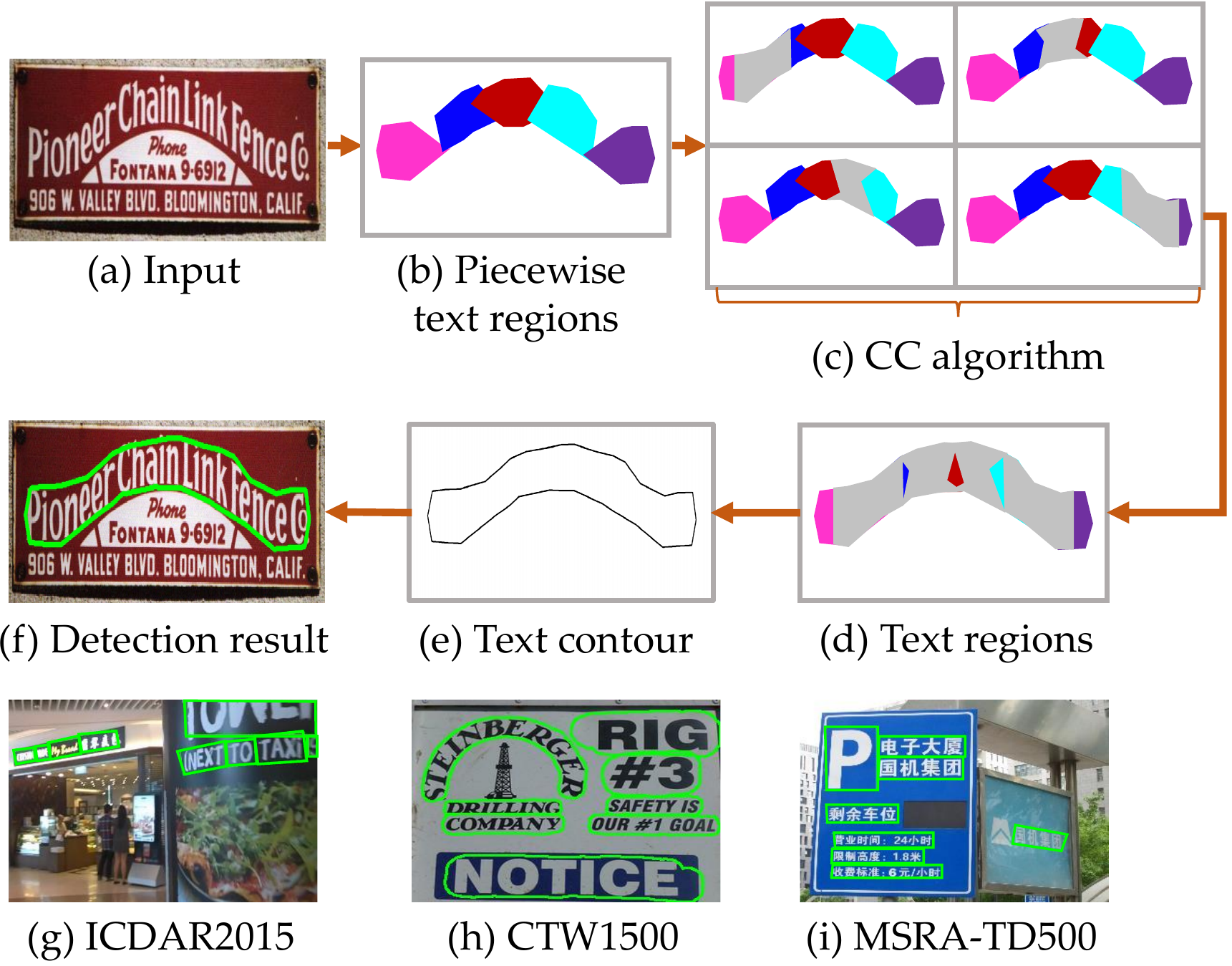}
	\end{center}
	\vspace{-0.5cm}
	\caption{Visualization the process of CC algorithm and some quality results on three public datasets.}
	\label{V9}
\end{figure} 
\subsection{Comparison with State-of-the-Art Methods}
In this section, we compare the proposed BiP-Net with previous methods on MSRATD500, CTW1500, and ICDAR2015 datasets to demonstrate the superiority of our method.

As we can see from Table~\ref{comparison}, our method achieves better comprehensive performance than other methods on all three datasets. Specifically, BiP-Net achieves 86.4\% in F-measure and 30.7 in FPS on MSRA-TD500, which surpasses the PAN~\cite{wang2019efficient} in detection accuracy and brings more than 20 FPS improvements compared with CRAFT~\cite{baek2019character} and SAE~\cite{tian2019learning}. The results demonstrate the superior performance of our method to detect long multi-lingual text instances. The same conclusion also can be obtained on CTW1500, where BiP-Net outperforms TextSnake~\cite{long2018textsnake} and TextRay~\cite{wang2020textray} by 9.5\% and 3.5\% about F-measure. For ContourNet~\cite{wang2020contournet}, BiP-Net not only outperforms it in F-measure but runs 7 times faster than it in the inference stage. The above experiments verify the effectiveness of the proposed text representation method on fitting irregular-shaped text contours. We visualize the inference process in Fig.~\ref{V9}(a)--(f). Moreover, on ICDAR2015, our method brings 0.8\% and 17.7 improvements in F-measure and FPS compared with the other best method, which proves that BiP-Net can detect small-scale and multi-oriented text instances effectively. Some quality detection results on three datasets are visualized in Fig.~\ref{V9}(g)--(i).

\begin{table}[]
	\footnotesize
	\setlength{\tabcolsep}{1.5mm}
	\caption{Time consumption of BiP-Net on three public datasets. ``Size'', ``Head'' and ``Post'' represent the size of short side of input image, sample center point and predict ray modules, and post-processing respectively.}
	\centering
	\renewcommand\arraystretch{0.8}
	\label{speed}
	\begin{tabular}{c|c|ccc|cc}
		\toprule
		\multirow{2}{*}{Dataset} & \multirow{2}{*}{Size} & \multicolumn{3}{c|}{Time consumption (ms)} & \multirow{2}{*}{F-measure } & \multirow{2}{*}{FPS} \\ \cmidrule{3-5}
		&                       & Backbone        & Head        & Post       &                            &                      \\ \midrule
		MSRA-TD500               & 736                   & 11.9            & 15.8         & 4.9        & 86.4                       & 30.7                 \\ \midrule
		CTW1500              & 640                   & 9.7              & 12.3         & 6.1        & 85.1                       & 35.6                 \\ \midrule
		ICDAR2015                & 736                   & 14.2            & 19.0        & 7.1        & 83.9                       & 24.8                 \\ \bottomrule
	\end{tabular}
\end{table}

\subsection{Ablation Study} 
We specially analyze the time consumption of BiP-Net in different stages
on three datasets. As we can see from Table~\ref{speed}, the head of the network takes the most time, and the time cost of post-processing is only about 20\% of the overall time consumption, which verifies the high efficiency of the proposed multiple ray clusters based text representation strategy for fitting arbitrary-shaped text contours. At the same time, by comparing the results that are conducted on ICDAR2015 and CTW1500, we can find that the time consumption of the BiP-Net network mainly depends on the image scales, which means our method can run faster for smaller size images.

\section{Conclusion}
\label{conclusion}
In this paper, we propose a bidirectional perspective network (BiP-Net) for arbitrary-shape text detection. A top-down perspective text representation strategy is proposed to model text instances with a fixed number of ray clusters. Compared with previous strategies, the proposed strategy can fit highly curved text contours more effectively. Moreover, a Contour Connecting (CC) algorithm is proposed to rebuild text contour from a bottom-up perspective to avoid losing text contours information. With the aforementioned advantages, BiP-Net achieves excellent performance in both detection speed and accuracy on three public datasets. Comparison experiments show that our method achieves a better balance between detection accuracy and speed compared with existing SOTA methods. Moreover, we analyze the time consumption of BiP-Net in different stages and the experiment results demonstrate the proposed text representation method can fit arbitrary-shaped text contours efficiently.


\vfill\pagebreak
\bibliographystyle{IEEEbib}
\bibliography{strings,egbib}

\end{document}